\documentclass[journal,transmag]{IEEEtran}
\usepackage{graphicx}
\usepackage{graphics}
\usepackage{subfigure}
\usepackage{calligra}

\usepackage{mathalfa}
\usepackage{amsmath}
\usepackage{amsfonts}
\usepackage{amssymb}
\usepackage{amsmath}
\usepackage{fancyhdr}

\ifCLASSINFOpdf
\else
\fi
\begin{document}
\title{Deep network as memory space: complexity, generalization, disentangled representation and interpretability}


%Montgomery Scott\IEEEauthorrefmark{3}, and
%Eldon Tyrell\IEEEauthorrefmark{4},~\IEEEmembership{Fellow,~IEEE}}
%\IEEEauthorblockA{\IEEEauthorrefmark{1}School of Electrical and Computer Engineering,
%Georgia Institute of Technology, Atlanta, GA 30332 USA}
%\IEEEauthorblockA{\IEEEauthorrefmark{2}Twentieth Century Fox, Springfield, USA}
%\IEEEauthorblockA{\IEEEauthorrefmark{3}Starfleet Academy, San Francisco, CA 96678 USA}
%\IEEEauthorblockA{\IEEEauthorrefmark{4}Tyrell Inc., 123 Replicant Street, Los Angeles, CA 90210 USA}% <-this % stops an unwanted space
%\thanks{Manuscript received December 1, 2012; revised August 26, 2015.
%Corresponding author: M. Shell (email: http://www.michaelshell.org/contact.html).}}

\author{\IEEEauthorblockN{Xiao Dong, Ling Zhou}
\IEEEauthorblockA{Faculty of Computer Science and Engineering, Southeast University, Nanjing, China}}

%{Shell \MakeLowercase{\textit{et al.}}: Bare Demo of IEEEtran.cls for IEEE Transactions on Magnetics Journals}



% for Transactions on Magnetics papers, we must declare the abstract and
% index terms PRIOR to the title within the \IEEEtitleabstractindextext
% IEEEtran command as these need to go into the title area created by
\IEEEtitleabstractindextext{%
\begin{abstract}
By bridging deep networks and physics, the programme of geometrization of deep networks was proposed as a framework for the interpretability of deep learning systems. Following this programme we can apply two key ideas of physics, the geometrization of physics and the least action principle, on deep networks and deliver a new picture of deep networks: deep networks as memory space of information, where the capacity, robustness and efficiency of the memory are closely related with the complexity, generalization and disentanglement of deep networks. The key components of this understanding include:(1) a Fisher metric based formulation of the network complexity; (2)the least action (complexity=action) principle on deep networks and (3)the geometry built on deep network configurations. We will show how this picture will bring us a new understanding of the interpretability of deep learning systems.

\end{abstract}

\begin{IEEEkeywords}
deep networks, geometrization, interpretability, geometrization of physics, quantum information, Riemannian geometry, complexity
\end{IEEEkeywords}}

\maketitle

%\affiliation{}

%\tableofcontents

\IEEEdisplaynontitleabstractindextext

% \fi
\IEEEpeerreviewmaketitle

\section{Introduction}

Till now the interpretability of deep learning systems remains the dark cloud above the sky of deep learning. Although we have developed useful tools and methods to understand deep learning systems, for example from the ordinary differential equation and the optimal control perspectives, still we are lacking of a general framework to describe and understand deep learning systems. For the interpretability of deep learning systems, not only we need to find out how deep networks process data to accomplish a specific task, also we need to answer why deep networks work like that. To make this point clearer, we can first examine some examples in physics. We all know spacetime is described by Einstein's general relativity, but till today we can not understand why spacetime obeys the gravitational equation. We know exactly how electrons interact with EM waves, but we still can not answer why we have electrons and EM waves. We believe the goal of the interpretability of deep networks should also be to answer those \emph{why} questions instead of those \emph{how} questions. So we need a profound and primitive principle to derive a mathematical framework to interpret deep networks and deep learning systems, or in another word, we need a programme for the interpretability of deep learning systems. From this point of view, the ordinary differential equation (ODE) or the optimal control perspective are more tools rather than the programme we ask for. The goal of this paper is to give our first attempt to construct the programme.

In our former work we proposed to understand the interpretability of deep learning systems by geometrization\cite{Dong2019geo}\cite{Dong2019over}. Obviously this idea is adapted from the geometrization of physics. Here we formally formulate our idea under the name of \emph{the programme of geometrization of deep networks}. We expect it can play a similar key role in the field of deep learning as its counterpart does in physics.

The motivation of the proposed framework stems from our belief that the universal effectiveness of deep learning in different fields must have a simple fundamental origin: \emph{deep network is physical}. The meaning of this slogan is two-folds. On one hand, deep networks are effective descriptors for our physical world so that all physical systems can be effectively described by deep networks. On the other hand, the physical world emerges from deep networks, to be more specific, from the deep networks of quantum computation. For every physical system, there exists a correspondent deep network to generate and describe it. While for every deep network, it also possesses a physical picture. So in this programme, physics and deep networks are intertwined and deep networks of quantum computation provide a constructive realization of Wheeler's idea \emph{It from qubits}, which claims that the physical world emerges from information. With this belief in mind, the goal of our programme of geometrization of deep networks is to bridge the concepts in physics and deep learning systems so that both fields can benefit from this programme. This is to say, in one direction, the idea that the physical world is emergent from quantum computation networks can help us to understand the fundamental structure of our universe such as spacetime, material and their interactions. In the other direction, key concepts and methods in physics such as the least action principles and geometrization of physics can be transferred to the field of deep learning to provide an interpretation of deep networks. We hope such a bi-directional win-win pattern can help to answer those \emph{why} questions in both fields and achieve a better interpretability of our world.

Indeed in the physics side, especially in the field of quantum information and quantum gravity, there are already some interesting results by considering how our physical world can emerge from quantum information processing procedure. Concepts such as computation, information and quantum complexity are now playing a more important role in understanding the fundamental rules of physics\cite{Lloyd2006A}\cite{Swingle2012Constructing}\cite{Raamsdonk2010Building}\cite{Matsueda2014Derivation}\cite{Wen2003Quantum}. One interesting observation is that there exist a geometry/information duality, which means that spacetime emerges from the information of a physical system and the emergent spacetime can be regarded as a memory space for that information\cite{Lloyd2012A}\cite{Evenbly2011Tensor}. In this paper, we will try to adapt this idea to deep networks so that deep networks can be understood as a memory device of the information represented by data, or, the geometry of a deep network encodes data information. This idea can be mathematically formulated by combining a Fisher information metric based definition of deep network complexity and the least action (complexity) principle on the configuration of deep networks. We will show that deep network as the memory of information, its capacity, robustness and efficiency are closely related with the complexity, generalization and disentangled representation of the deep network. We claim that these observations can bring us new understanding of the interpretability of deep learning systems.

The remaining part of this paper will be arranged as follows. We will first give a brief summary of the key idea of our programme of geometrization of deep networks and the geometry/information duality in physics. Then we will explain how this idea can help us to understand the key concepts in deep learning systems such as the network complexity, generalization performance and disentangled representation. For the same perspective, we will also give a discussion of some interesting recent works, including the Lottery hypothesis, the weight agnostic neural network, the disentangle representation and the information based interpretation of NLP systems.

\section{Geometry/information duality in physics}

Geometrization of physics is one of the most profound and the most successful idea in understanding the rules of our physical world in human history. But \emph{why }can our world be geometrized? In the last decade, we saw a new trend to combine geometrization and quantum information processing to scratch a complete new picture of our world. Basically this is to regard our world, including spacetime, material and their interactions, as emergent phenomena from a complex deep network of quantum computation, where the deep network aims to represent or generate the quantum state or the information of the universe. This is the key idea of the so called computational universe. From this point of view, our world is built from deep networks and the geometric structure of the physical world emerges from the geometric structure of the underlying deep networks. So the geometrization of physics \emph{is} essentially the geometrization of the underlying quantum deep networks. The success of geometrization of physics is a hint that geometrization may also be the key to understand deep networks. This is the motivation of our proposed programme of geometrization of deep networks. We wish this can serve as a general framework for the interpretability of deep learning systems.

In computational universe, there is a deep connection between the information of the universe and the geometrization of the universe, which are connected by a complex quantum computation deep network\cite{Lloyd2012A}\cite{Evenbly2011Tensor}. In this section we will explore the duality between geometric structures and information, which means that in our physical world, geometric structures emerge from information and geometric structures encode information.

The basic logic of this geometry/information duality can be summarized as follows.

1. Firstly, for a given quantum system, its quantum state $|\psi\rangle$ has an information pattern, which includes, for example, the entanglement entropy and correlation among subsystems.
2. We may also have different descriptions $D(|\psi\rangle)$ of this quantum state $|\psi\rangle$, for example either by a tensor network\cite{Evenbly2011Tensor}\cite{Evenbly2017Algorithms} or by a quantum circuit\cite{Nielsen_geometry2} which can be used to generate $|\psi\rangle$ from a simple reference quantum state such as a product state.
3. By assigning a proper definition of complexity $C(D(|\psi\rangle))$ of the descriptions of $|\psi\rangle$, we can apply the least action principle on the descriptions to find the optimal description of $|\psi\rangle$, where the action is given by the complexity following Susskind's idea that Complexity=Action\cite{Susskind2017Dear}.
4. And finally, the optimal descriptor of $|\psi\rangle$ holds a geometry that encodes the information pattern of $|\psi\rangle$\cite{Evenbly2011Tensor}, which can be called the memory space.  %For example, in quantum information field, a MERA, which can be understood either as a tensor network or a quantum computation program, based description of a quantum state holds an Anti de Sitter geometry and it encodes the information pattern of the correspondent quantum state. In quantum gravity, quantum computation based spacetime and the information geometry based derivation of the gravitational equation confirm that our spacetime goemetry emerges from the quantum states of our universe and it also encodes the information of our universe.%

In the above mentioned steps, we in fact encounter two geometric structures.

The first geometric structure is the Riemannian manifold of all descriptions of all quantum states. A description $D(|\psi\rangle)$ is a point on this manifold and its complexity $C(D(|\psi\rangle))$ is defined as the geodesic distance between $C(D(|\psi\rangle))$ and the trivial description $I$. We call it the geometry of complexity $G_{Com}$. One example of this Riemannian geometry is the geometry of the quantum circuit complexity of quantum computation defined in \cite{Nielsen_geometry2}. In our former work on the geometry of deep networks\cite{Dong_deep}\cite{Dong2019geo}, we have shown the Riemannian structure of quantum computational complexity already provides a geometric picture for deep networks, which is also related with geometric mechanics as in the diffeomorphic image registration problem\cite{Bruveris2013Geometry}. Here we point out that this geometric structure is in fact compatible with the recent ODE and optimal control based picture of deep networks since the diffeomorphic image registration problem is essentially be formulated as both an ordinary differential equation and an optimal control procedure\cite{Weinan2018A}. From this observation, the ODE or optimal control picture of deep networks also explores the geometrization of deep networks.

Besides the Riemannian geometry of quantum complexity, the optimal descriptor of a quantum state may also generate a geometric structure called the geometry of information $G_{Inf}$. For example a tensor network as a descriptor of quantum states may have a physical geometry or a holographic geometry as in \cite{Evenbly2011Tensor}. Another example is the theory of quantum computation based gravity\cite{Lloyd2006A}\cite{Lloyd2012A}, where spacetime geometry emerges from histories of quantum computation. The key concept of the second geometric structure is that the emergent geometry can encode the information pattern of the quantum state $|\psi\rangle$. In another word, the geometric structure is a memory to save the quantum information of $|\psi\rangle$. For example, MERA can be regarded as a discrete realizations of the AdS/CFT duality and its holographic geometry encodes the information pattern of the boundary quantum states such as the area law of entanglement entropy and the correlation length\cite{Evenbly2011Tensor}\cite{Evenbly2017Algorithms}\cite{Swingle2012Constructing}.

As a summary, in physics, we have now a correspondence or duality between information and geometry. The duality is established by a least action principle where the action is given by the complexity of the quantum state, or equivalently the complexity of the optimal descriptor of the quantum state.

Given this picture, a set of straight forward questions pop up. What's the relationship between these two geometric structures? If the second geometry $G_{Inf}$ is a memory to encode information, how can we evaluate its capacity, robustness/reliability and efficiency? Can we somehow unify these two geometric structures? Because these two geometric structures are related with the metric of quantum complexity and information respectively, a natural guess is that, they might be unified by the Fisher information metric. In fact, it seems they do! This means that in $G_{Com}$ the proper complexity metric should be the information metric and it can also be used to evaluate the performance of the geometric memory of information $G_{Inf}$.

An intuitive understanding of this point is that information metric evaluates how the information changes with respect to the configuration parameters of an information processing system. And in our discussion of the information/geometry duality, all we are interested is just to describe the quantum state $|\psi\rangle$ with a quantum information processing system, either a tensor work or a quantum computation algorithm. In fact the Fisher information metric has been used to derive the gravitation equation and the AdS/CFT correspondence. This is a strong evidence that information metric may play a key role in understanding deep networks following the programme of geometrization of deep networks. We hope this can give a richer geometric picture of deep networks and help to clarify some important issues in deep learning systems.

\section{Geometry/information duality in deep networks}

Similar to the information/geometry duality in physics, we can also build two geometric structures of deep networks and examine the geometry/information duality of deep networks. Conceptually this is straight forward since as an information processing system, the deep network of quantum computation is more fundamental and its geometry/information duality can be directly applied to our classical deep networks. We formalize this procedure as follows.

For a given task $\mathbf{T}$ with its training data set $\mathbf{X}$, we build a $L$-layered deep network with its structure and parameter set denoted by a graph $G$ and $\mathbf{\theta}$ respectively. This gives a simplified description of deep networks.

\textbf{A. Information in deep networks}\\
To build the geometry/information duality of deep networks, we need first to clarify what's the \emph{information} here. Obviously the information is now encoded in data $\mathbf{X}$. Depending on the task $\mathbf{T}$ of the deep learning system, the information can be the distribution $f(\mathbf{X})$ of data $\mathbf{X}$ if we are building a generative model for data $\mathbf{X}$, or it can be a pattern classification of data $\mathbf{X}$ if we are working with an image classification system. In these examples, the information to be saved in the deep network is the different patterns of data or the structure of the error-correcting code since generator of a GAN is a cascaded error-correcting code. We can assume the information is represented by a set of random variables $\mathbf{Y}$ and this information should be recorded by a deep network.

\textbf{B. $G_{Com}$ of deep networks}\\
As an efficient tool to approximate complex functions, deep networks can be understood as discretized ordinary differential equations. From this ODE perspective of deep networks, each deep network is a state flow determined by the correspondent ODE and it works as a continuous transformation flow that transforms the data from input to output of the deep network. Obviously on the manifold of all functions that can be approximated by deep networks, each deep network corresponds to a continuous curve. This is exactly the same as in quantum computation, where a quantum computation algorithm is a continuous quantum evolution path and the quantum circuit based realization of a quantum algorithm is a discretized curve\cite{Nielsen_geometry2}. Another analogue is the diffeomorphic image registration problem, where each image transformation trajectory is a continuous curve. Our former work has shown that these three aforementioned systems have exactly the same Riemannian structure. For more analysis of $G_{Com}$, please refer to \cite{Dong2019geo}\cite{Dong_deep}.

What's the role of the deep network structure $G$ and the parameter $\mathbf{\theta}$ here? Obviously, $G$ defines a submanifold of functions that can be represented by deep networks since a given network structure $G$ can only approximate a subset of all the functions on this manifold. Different configurations of $\mathbf{\theta}$ will then give different curves on this manifold determined by $G$. So training a deep network is to find a curve on this manifold to reach a target function that can accomplish a certain task.

\begin{figure}
  \centering
  \includegraphics[width=8.5cm]{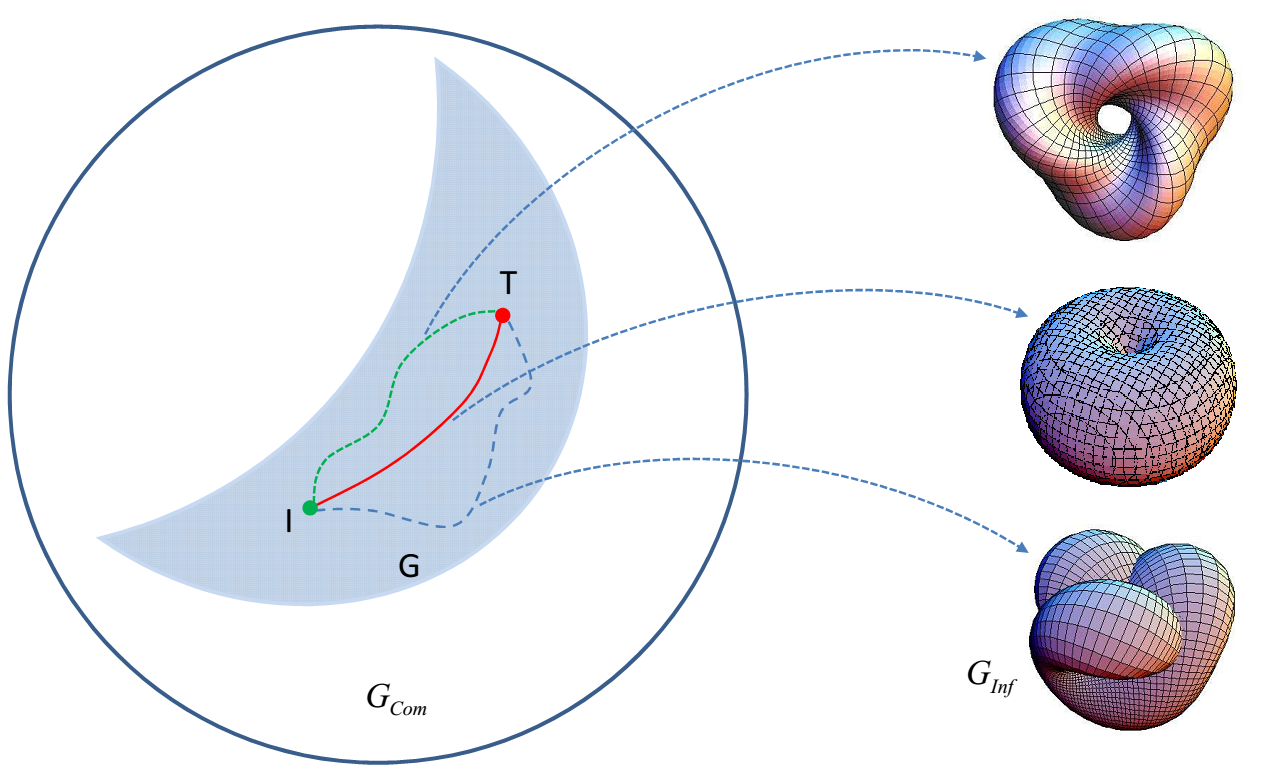}
  \caption{Geometry of deep networks. $G_{Com}$ is a Riemannian manifold of the functions that can be represented by deep network, where the network structure $G$ defines a submanifold and different network configurations with the same structure $G$ correspond to different curves on $G_{Com}$ connecting the trivial identity function $I$ and the target function $T$. Different deep networks also have their correspondent emergent geometric structures $G_{Inf}$, which can encode the information of data so that deep networks can be regarded as a memory space of information.}\label{fig-1}
\end{figure}

The geometric structure $G_{Com}$ is then the Riemannian structure defined on the manifold of all deep networks, or the manifold of all the functions that can be achieved by deep networks. The Riemannian metric on this manifold is to define the arc length of the curves on the manifold, or more precisely, an inner product on the tangent space of the manifold. Given this Riemannian metric, we can then define the complexity of a deep network, a curve on this manifold, as the length of its correspondent curve. The introduction of the Riemannian metric enables us to discriminate different deep networks even they represent exactly the same target function since they may have different complexities. We will see this is a key concept for the interpretability of deep networks.

Here we would like to clarify two problems. \\
(1) How big is the manifold $G_{Com}$?\\
Following the universal approximation rule of neural networks, some may think deep networks can approximate any functions. But this is not true. The key property of deep networks is that they are effective descriptors for our physical world. When we are using deep networks to solve problems, we are only seeking for efficient solutions. In physics, a generic n-qubit quantum system in a state $|\psi\rangle$ has a quantum state complexity of $O(2^{2n})$ and therefore it can not be efficiently described or prepared. Only a small subset of simple quantum states is physically realizable and can be described efficiently, which is called \emph{the corner of physical world}\cite{Or2014A}. Similarly, not all functions can be efficiently approximated by a deep network with a limited size. We can only work with a subset of all functions, which we can denote as \emph{the corner of physical functions}. In quantum computation, we have the same problem. When we are working on a quantum information processing system with $n$ qubits, a general algorithm can be described by a unitary operator $U \in U(2^n)$. But a general $U \in U(2^n)$ has a complexity of $2^{2n})$ and we are only interested in the algorithms that can be efficiently achieved by a limited number $N$ of simple quantum gates, for example a sequence of one and two qubit quantum gates with $N<<2^n$. This is exactly we are more interested in deep and narrow networks instead of shallow and extremely wide networks.  Physics will never accept arguments that can not be physically verified and we believe this also applies to deep networks.

(2) Which Riemannian metric should we choose to measure the complexity of deep network?\\
In recent years, complexity is becoming a key concept in physics including quantum information processing, quantum gravity and quantum phase transition. Susskind's Complexity=Action and Complexity=Volume hypothesis turn computational complexity into a concrete physical concept with physical correspondence\cite{Susskind_ER_bridge}\cite{Susskind2017Dear}. In our former work we also proposed to connect quantum complexity with space and time structure. According to the close connection between deep networks and physics, complexity should also be a key concept in deep learning systems.

If we agree that the complexity of a deep network is given by the length of the correspondent curve on the manifold of the corner of physical functions, then basically the complexity can be defined by either an energy based metric or an information based metric. The energy base metric aims to define the complexity by finding out how much energy should be consumed to carry out the information processing task on a deep network. The quantum circuit complexity of Nielsen is one example of it. In deep learning systems, complexities defined on the norm of the network parameter $\mathbf{\theta}$, for example the 0-norm to count the number of parameters or the 2-norm to compute the energy of the parameters. On the contrary, the information based metric also considers how the network configuration will influence the information representation of the deep network. Intuitively it concerns not only how much effort we should pay to accomplish the information processing task, but also how our effort can influence the output effectively since different operations with the same energy consumption may lead to different effects on the information processing. In physic, Fisher information metric is commonly used as an information based metric. It's well known that Fisher information metric provides a natural Riemannian metric to justify how the parameter of an information processing system will influence information distribution. In fact this information metric has been used to examine the AdS/CFT correspondence and gravity\cite{Matsueda2013Emergent}. This is a convincing evidence that information metric might be the optimal complexity metric for deep network complexity. In \cite{Bruveris2016Geometry}, it was proved that on a closed manifold of dimension greater than one, every smooth weak Riemannian metric on the space of smooth positive probability densities, that is invariant under the action of the diffeomorphism group, is a multiple of the Fisher–Rao metric. This diffeomorphism invariant property perfectly matches the diffeomorphism invariance of spacetime. This is highly desirable for both the complexity of deep networks and the geometry $G_{Inf}$ which is essentially a relational structure.  Besides these, in \cite{pmlr-v70-koh17a}\cite{pmlr-v80-ren18a} Fisher information metric was used to analysis the influence of data, and in \cite{Liang2017Fisher}\cite{Lei2017Towards}\cite{Karakida2018Universal} the Fisher information metric based network complexity was used to evaluate the generalization property of deep networks.  From all these former works, we believe Fisher information metric should be the optimal metric for deep network complexity.

If the Fisher information metric is chosen to define network complexity, it leads to a way to evaluate how the network parameter $\mathbf{\theta}$ will influence the information encoded in the network. Assuming the information saved by the deep network is the distribution of a random variable $y$ with a probability distribution $p(y;\mathbf{\theta})$, then we have
\begin{equation}\label{eq-1}
\begin{split}
  D_{KL}&\approx \frac{1}{2}g_{\mu\nu}(\mathbf{\theta})d\theta^\mu d\theta^\nu \\
  \gamma(y;\mathbf{\theta})&= -\ln(p(y;\mathbf{\theta}))\\
  g_{\mu\nu}(\mathbf{\theta})&=\int p(y;\mathbf{\theta})\frac{d\gamma(y;\mathbf{\theta})}{d\theta^\mu}\frac{d\gamma(y;\mathbf{\theta})}{d\theta^\nu}dx=<\partial_\mu\gamma\partial_\nu\gamma>
\end{split}
\end{equation}

where $D_{KL}$ is the Kullback-Leibler divergence of two distributions.

Obviously the information of $Y$ is saved in a spacetime with coordinates $\mathbf{\theta}$ where the distance is measured by the Fisher metric. The matrix $g_{\mu\nu}(\mathbf{\theta})$ encodes how robust the information of $y$ with respect to the network parameters $\mathbf{\theta}$. An intuitive picture of this is that, a larger $g_{\mu\nu}(\mathbf{\theta})$ means the information encoded in the network is more sensitive to the network parameters $\theta^\mu, \theta^\nu$\cite{Matsueda2013Emergent}.

C. \textbf{$G_{Inf}$ of deep networks}\\
Following the same logic of the geometry/information duality in physics, we can apply a least complexity principle on the network configuration and the network training will result in a low complexity deep network, on which $G_{Inf}$ will emerge. In order to understand the structure of $G_{Inf}$, we need to recap how the $G_{Inf}$ can be built in physics.

The first example is spacetime from quantum computation \cite{Lloyd2012A}. According to this theory, spacetime emerges from a deep network of quantum computation. This is an example of a purely relational theory since spacetime jumps out of a history of quantum computation without any background geometry. So this is a background independent construction of $G_{Inf}$. The building blocks of spacetime are the quantum operations so that each quantum gate in the deep network can be regarded as a point with an unknown volume or an event in spacetime. Since we are working with a background independent scenario, all points are floating in nowhere before $G_{Inf}$ is constructed. The task of building a geometric structure $G_{Inf}$ is to embed all points in a geometry so that the volume, area and distances between points can be determined. In quantum computation based spacetime, the geometry is built by a variational principle so that an action is minimized with respect to the geometry\cite{Lloyd2012A}. There the action is defined on both the curvature of the geometry and the operations in this quantum deep network. This is why the gravitational equation can be reproduced, which connects both the geometry of spacetime and materials. We will not further dive into further details of this procedure, instead we only summarize some interesting observations in this example.

(1) The geometry is built on the operations of the quantum computation procedure, or the computation nodes of the deep network. The computational nodes can be understood as events in spacetime.\\
(2) This is a background independent construction of a geometric structure so that the geometry is completely determined by the relations between nodes. This is to say, we are working with a completely relational theory. The dimension of the emergent geometry is determined by the dimension of the quantum operation. In the quantum computation based spacetime, the quantum operation is a 2-qubit gate with a spectrum of dimension of 4. This is why we have a 4-dimensional spacetime.\\
(3) The resulting geometry, the spacetime with the gravitational equation, is diffeomorphism invariant. \\
(4) The emergent geometry, the curvature of the Riemannian structure, encodes the information of the quantum deep network, which is the history of the quantum computation carried out by the deep network.\\
(5) The geometry is a result of a statistical average of the histories of all data passing through the quantum computation network. This is because depending on the input quantum states, the quantum deep network will have different histories of quantum operations. This is the same as in a normal deep network, different input data will have different patterns of information propagation in the deep network so that the geometry $G_{Inf}$ is a result of the statistical average of all the data $\mathbf{X}$. This is exactly why the Fisher information metric $g_{\mu\nu}(\mathbf{\theta})$ is defined by a statistical average in \ref{eq-1}.\\

Another example of the emergent geometry in physics is the geometry of tensor network states, where tensor networks are used to represent different quantum states\cite{Evenbly2011Tensor}. Physical or holographic geometries can be built on tensor networks and the geometry of a tensor network encodes the information pattern of the correspondent quantum states, for example its quantum entanglement and correlation between subsystems. For more details of the geometry of tensor networks, please refer to \cite{Evenbly2011Tensor}. Similar to the spacetime structure from quantum computation example, we only give a brief summary of the $G_{Inf}$ as follows.

(1) The geometry is built on the tensors of the tensor networks, which can also be regarded as deep networks.\\
(2) This is a background dependent construction of geometric structures so that the emergent geometry depends on the dimension of the background geometry.\\
(3) The tensor network  encodes the information of the quantum state in its geometry.\\
(4) The geometry of a tensor network is also statistical since it encodes the information pattern of a set of quantum states in the same complexity class. \\

The aforementioned two different $G_{Inf}$s share some common characteristics and also they differ from each other. The picture of our $G_{Inf}$ of deep networks will be a combination of these two pictures.

Before we give the picture of $G_{Inf}$ of general deep networks, we would like to have a short discussion of the geometry of spacetime since the aforementioned two special $G_{Inf}$s are both spacetime geometries, i.e. the spacetime of our universe and the holographic spacetime respectively. We will see some of their structural properties discussed here will also appear in $G_{Inf}$ of deep networks.

The first question is, in the quantum computation based spacetime, even the spectrum of the 2-qubit gate determines the emergent spacetime is four dimensional, why do we have a three dimensional space and a one dimensional time? Shouldn't the 4 dimensions of the spectrum of 2-qubit gates stand on the same foot? From the quantum information point of view, time is related with the evolution of quantum complexity\cite{Susskind2016The}\cite{Susskind_ER_bridge_nowhere}. The reason that we have a thermodynamic arrow of time is that statistically the quantum complexity of a quantum system can only increase. Similarly, along the computation of the quantum deep network the quantum complexity of data will statistically increase, so we have time in the deep network of quantum computation. But still this can not explain how the 4 dimensions diverges into a three dimensional space and a one dimension time. From the geometry/information duality perspective, spacetime is a memory to save the information of the quantum deep network, i.e. the quantum algorithm carried out by the deep network. At each quantum operation node, the 4 dimensional variables $(\theta_1,\theta_2,\theta_3,\theta_4)$ encode part of the information of the quantum algorithm denoted as $y$. If we regard the quantum algorithm as a random variable, then its distribution can be written as $p(y;\theta_1,\theta_2,\theta_3,\theta_4)$ so that the information of $y$ depends on $(\theta_1,\theta_2,\theta_3,\theta_4)$. According to \ref{eq-1}, we can easily see that the information of $y$ has different sensitivities on different quantum computation nodes. That's to say, for different 2-qubit gates inside the quantum deep network, the variations of different gates will have different effects on the information of $y$. From the knowledge of the complexity of quantum computation\cite{Nielsen_geometry2}, statistically the sensitivity smoothly changes along the layers of the quantum deep network so that the information of $y$ has a higher sensitivity to the nodes at the lower layers. So $(\theta_1,\theta_2,\theta_3,\theta_4)$ can be normalized as $(\theta_1,\theta_2,\theta_3,\theta_4)=s(\theta_1',\theta_2',\theta_3',\theta_4')$ with $s$ related with the sensitivity of the current node. After this normalization, the information of $y$ has roughly an equal sensitivity on all the nodes except for a variable scaling factor $s$ along the network. The variable scaling factor $s$ can then be regarded as time. Due to the renormalization operation, $(\theta_1',\theta_2',\theta_3',\theta_4')$ is essentially only 3 dimensional, which corresponds to the 3 dimensional space. In fact the observation that time corresponds to the sensitivity also appeared in the work of Susskind \cite{}, which aims to address the spacetime structure of black holes.

Another related question is about the Planck length $l_{p}$ in our spacetime, which is the minimum length that we can distinguish. Why do we have $l_p$? In fact we can give different pictures about the existence of $l_p$ from different perspectives. From the uncertainty principle of quantum mechanics, $l_p$ exists since we need a minimal time $t_p$ to evolve an initial quantum state to a final quantum state that is orthogonal/distinguishable to the initial state. From the geometric point of view, $l_p$ exists since the spacetime can not curve to more than it can curve\cite{Lloyd2012A}. Finally from the geometry/information duality point of view, since geometry encodes information, the minimal length $l_p$ means that a too small change of the geometry will not significantly change the information it encodes. Or equivalently the geometry is determined by $\mathbf{\theta}$, if we regard $\mathbf{\theta}$ as an unknown parameter that should be estimated from the information (the quantum state) of our universe, the Fisher-Rao limit of this parameter estimation will result in $l_p$ so that the minimal variance of the estimate of $\mathbf{\theta}$ corresponds to a perturbation of the spatial distance given by $l_p$. So $l_p$ is a signature to indicate how robust the geometry can save the information of our universe. A larger $l_p$ means a more robust memory. Immediately we can ask, how about black holes? They are abnormal spacetime structures whose internal details can not be detected for observers outside the event horizon. Intuitively the radius of a black hole corresponds to the Fisher-Rao bound that an outside observer can estimate the interior geometry of the black hole. So black holes are robust memories of quantum state information and the geometry of a black hole is not sensitive to the details of the quantum state of the material that collapsed to the black hole. This is exactly the no-hair theorem of black holes. Intuitively, a robust memory space means a smaller $g_{\mu\nu}(\mathbf{\theta})$ so that a change of the parameter $\mathbf{\theta}$ will lead to a minor change of information. Immediately we can see, a robust memory space corresponds to a deep network with a smaller complexity. We will see this is the key concept for the generalization property of deep networks.

Given the pictures of n $G_{Inf}$ in different physical systems, we can now try to scratch a global picture of $G_{Inf}$ for general deep networks as follows.
(1) The geometry is built on the computation nodes of deep networks, which are events of $G_{Inf}$.
(2) The emergent geometry can either be background independent or background dependent. In either case, the geometry is determined by the information and also it encodes the information of the deep network, i.e. the information of the computation carried out by the deep network.
(3) The robustness of the emergent memory space depends on the Fisher metric at different nodes. Generally speaking, a deep network with a lower complexity is more robust and this is also related with the generalization of the deep network.
(4) The emergent geometry is a statistical average on all the geometries of the data $\mathbf{X}$. This can be understood as that the geometry of spacetime can be constructed by collecting the trajectories of different free falling particles, where each individual data $x \in \mathbf{X}$ can generate a trajectory when it's processed by the deep network.

So how ca $G_{Inf}$ save the information? Intuitively $G_{Inf}$ is a complex landscape built from the computation nodes. A specific data can be regarded as a particle with a certain mass and an initial velocity, or a wave with a certain wave form. Then the data passes the network just as a particle or a wave passes a complex landscape $G_{Inf}$. Different data, as particles or wave forms, will lead to different trajectories or histories. This is very similar to the picture of how the wave function of a particle passes the spacetime in quantum field theory. So the information is saved in the geometry of $G_{Inf}$, which is used to guide the trajectories or the histories of different data.

\section{Deep networks as memory space of information}
Now we will examine how the geometry/information duality of deep networks can bring us new understanding of the interpretability of deep networks and deep learning systems. In our former work we already addressed the picture of $G_{Com}$ and we showed $G_{Com}$ shares the same Riemannian structure as quantum computation and geometric mechanics\cite{Dong2019geo}. In this paper we will focus on the situation that if the complexity metric is fixed as the Fisher information metric, what can we learn by taking deep networks as an information memory space. We will check the capacity, robustness and efficiency of the memory and we can see these issues are closely related with the complexity, convergence, generalization and disentangled representation of deep networks.

\textbf{A. Capacity, over-parameterization and general relativity}\\
The capacity of the memory is obviously related with the size of the deep network and a larger (deeper and wider) network has potentially a larger capacity. The widely used over-parameterized deep network is a memory with a high capacity. The observation of no bad local minima of over-parameterized network means that the high capacity memory can have different ways to save the same information by constructing different geometric structures.

The geometry of deep networks, including both $G_{Com}$ and $G_{Inf}$, depends on both the structure $G$ and the parameter $\mathbf{\theta}$. The network structure $G$ sets a prior constraint on the geometry and it has a strong impact on the geometry and also the dynamics of the network training. For example the success of ResNet largely depends on the fact that as a discretized time variant differential equation, the regular structure of ResNet defines a submanifold of deep networks with smooth geometries of $G_{Com}$. So the training of ResNet is in fact to find a curve on a smooth manifold and of course the convergence property of ResNet is superior to general deep networks.

Here we would like to point out that in quantum many-body systems, we have an analogue of over-parameterized deep networks, which is the MPS state with its correspondent parent Hamiltonian and uncle Hamiltonian\cite{Fern2015Frustration}. An MPS state is a low complexity quantum state on a 1-dimensional chain. It can be represented by a MPS tensor network and also can be represented as the ground state of a local Hamiltonian. If we compare the MPS states and over-parameterized deep networks, we will find
(1)The local Hamiltonian corresponds to the cost function of deep networks. Parent Hamiltonian and uncle Hamiltonian are non-overparameterized and over-parameterized networks respectively. The ground state of the local Hamiltonian is the configuration of the deep network which minimize the cost function.\\
(2)The MPS state is the solution of the deep networks that minimize the cost function. Or in the view of geometry/information duality, the MPS state is the optimal tensor network representation of the quantum state.\\
(3) The MPS state is an analytical solution of the network that minimize the cost function and we do not need to use gradient based optimization to find it.\\
(4) The MPS state shows a perfect geometry/information correspondence since the information represented by the Hamiltonian and the geometry represent by the MPS tensor network are analytically connected.
This shows that overparameterized deep networks also appear in physics and the comparison between overparameterized deep networks in both fields may provide an new approach to understand overparameterized networks, for example the geometry of the subspace of local minima of deep networks.

Another interesting observation is that when the Fisher information metric is used to define the network complexity, the resulting $G_{Inf}$ of the optimal network configuration and the information of data $\mathbf{X}$ show the same interaction pattern as the interaction between spacetime and material in general relativity. Following John Wheeler,\emph{spacetime tells matter how to move, matter tells spacetime how to curve}. In deep networks, \emph{network tells data (information) how to move, data (information) tells network how to curve.} This observation lies in the fact that due to geometry/information duality, both the spacetime structure and the optimal deep network configuration emerge from the same least action principle, where the action is given by the Fisher information metric based complexity, and therefor their dynamic rules share the same property. The similarity between deep networks and general relativity confirms our belief that our programme of geometrization of deep network is the right way to understand deep learning systems.

\textbf{B. Robustness, generalization and quantum mechanics}\\
The robustness of the memory is described by the Fisher information matrix, or the mean Hessian matrix as shown in \ref{eq-1}. In \cite{Lei2017Towards} it was shown that for an over-parameterized deep network, network training with a random initialization will converge to a configuration with a better generalization performance with a higher probability. They also assumed the configuration whose Hessian matrix has a spectrum with more small eigen values will have a larger attraction basin volume and also such configuration has a better generalization performance. But they gave no proof on this point. Also in \cite{Karakida2018Universal} the distribution of the spectrum of the Fisher information matrix was investigated to analyse the generalization performance of deep networks. 

Here we see the robustness issue provides an answer to the generalization problem of deep networks. For a network configuration whose Hessian matrix spectrum has a lot of small eigen values, obviously the information it encodes is less sensitive to the network parameter as shown in \ref{eq-1}. This means a small perturbation of the network configuration, equivalently the geometry of $G_{Inf}$ of the deep network, will not significantly change the information pattern that the network encodes. This is exactly what a good generalization asks for. Now we can answer why over-parameterized network seems always find a solution with a good generalization performance. The arguments can be formulated as follows:
(1) A network with a good generalization property means that it can save the information with a higher robustness. Or equivalently, a low complexity network has a better generalization capability. This is because the geometry $G_{Inf}$ of a low complexity network can be determined by a relatively smaller training data set so that it has a better generalization property .\\
(2) A high robustness means the information pattern is not sensitive to the perturbation on network configuration, so that the Fisher information matrix has a spectrum with more small eigen values.\\
(3) Such a network configuration has a larger attraction basin volume.\\
(4) Network training from a randomly initialization will fall in such a network configuration with a higher probability.\\

Here we would like to emphasize that\emph{with Fisher information metric, the network complexity is not proportional to the size of the deep network. So a deep network with a large number of parameters can still result in a low complexity network after training}. This observation may resolve the conflicts between the theoretical analysis of generalization bound and experimental results in PAC Bayes learning. This is to say, there are experimental works show that training a larger deep network results in a better generalization performance, which seems to contradict the PAC Bayes generalization bound. But if the network complexity is defined by the Fisher information metric, then a larger deep network may in fact have even a smaller network complexity after training, so that the superfacial contradiction may actually be resolved.

We can also find that the aforementioned observation that the probability that the network reaches a certain configuration from a random initialization is related with the complexity of that configuration\cite{Lei2017Towards} has a physical analogue. If we consider that network complexity is actually defined by a Riemannian distance, then we have a connection between the probability and a Riemannian distance. This is exactly what happens in the geometric formulation of quantum mechanics\cite{Heydari_dynamicdiatance}, where in quantum measurement the probability that a state collapses into a certain state is related with the distance between the initial and final states. Maybe this can bring us a new perspective to understand the quantum measurement problem. Again, this analogue is another evidence to support our programme of geometrization of deep networks.

\textbf{C. Efficiency, disentangled representations and quantum entanglement}
In the geometry/information duality, the least complexity principle means we are looking for a low complexity representation of the information so that the memory has a high efficiency. We believe this is also related the disentangled representation that deep learning systems are chasing for. From our geometrization point of view, the so called disentangled representation is just a natural byproduct of the least complexity oriented optimization. This is to say, the optimal low complexity deep network must give a disentangled representation of the information and the disentangled representation is just a way to achieve a low complexity.

Still we can find a counterpart in physics since the word entanglement is more popular in quantum computation. In quantum mechanics, the key feature of an entangled state is that its information is encoded in the global system but not in its subsystems. Accordingly disentanglement means information is saved in individual subsystems and we can play with the information just by manipulating subsystems. In fact this is exactly what the disentangled representation means in deep learning systems. In quantum information processing,  a quantum algorithm on n-qubits can be represented by an unitary operator $U \in SU(2^n)$ and its optimal implementation is to find a quantum circuit (a deep network of quantum computation) with a minimal complexity that can achieve $U$ by simple 1 and 2 qubit operators. We can immediately see the least complexity deep network for $U$ is exactly a disentangled representation of $U$ since all the 1 and 2 qubit operators only work on subsystems.

So network complexity is also a criteria for the disentangled representation. The recent work\cite{Locatello2018Challenging} claimed that in unsupervised learning, the disentangled representation can not be obtained without inductive bias on both the data and models. But if the nature of disentangled representation is a low complexity representation, to find it we must take the network complexity as part of the cost function. The assumption of \cite{Locatello2018Challenging} that we can not introduce any inductive bias in the unsupervised learning is meaningless, since it's not compatible with the essence of the disentangled representation. Different representations for the same information can be discriminated by their complexities. So in order to find disentangled representations, we must take the deep network complexity into consideration.

For more information on the geometric structure of disentangled representations, please refer to \cite{Dong2019gauge} where a fibre bundle based description of disentangled representations was introduced. There we see disentangled representations can be formulated and understood as a gauge theory, and the arguments of \cite{Locatello2018Challenging} were analysed in more details.

\begin{figure}
  \centering
  \includegraphics[width=8.5cm]{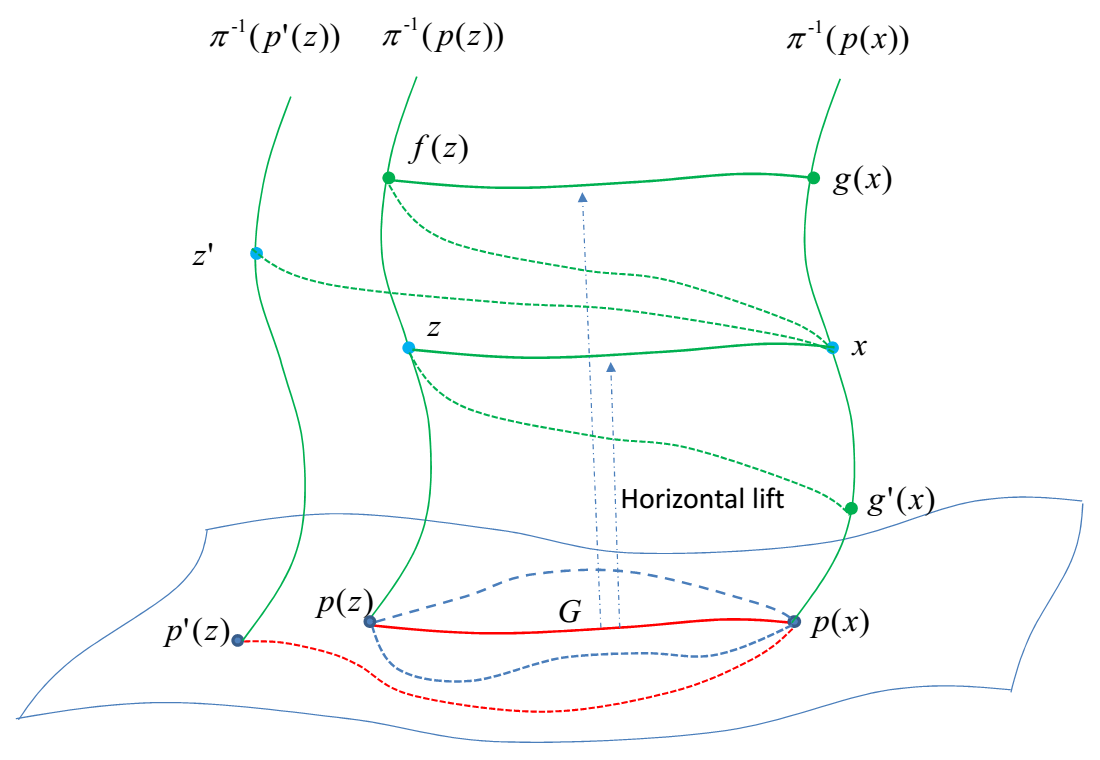}
  \caption{The fibre bundle structure of disentangled representations\cite{Dong2019gauge}}\label{fig-2}
\end{figure}

\textbf{D. Data augmentation, data distillation and data re-weighting}\\
How can $G_{Inf}$ encode the information in its geometry? According to the construction of $G_{Inf}$, the geometry is built on the deep network, where computation nodes are taking as points and the landscape of $G_{Inf}$ is given by the information metric. Obviously the information is stored in the landscape of $G_{Inf}$. An intuitive picture of how $G_{Inf}$ encodes the information is that:

The deep network constructs a complex landscape. An input data is a particle or a wave form incident to this landscape with a certain initial position and velocity. Along its travel across the landscape, its trajectory is determined by the landscape till it reaches its destination so that different particles or wave forms will have different trajectories. On the other hand, the landscape is shaped by the trajectories of all training data just as the spacetime is built by the histories of quantum computations\cite{Lloyd2012A}. So the whole story of deep learning is to build a landscape by adjusting the trajectories of training data so that the landscape can store the information of the correspondent task.

 Now we can check how the training data can shape $G_{Inf}$. There are three different ways to manipulate the training data, data augmentation, data distillation\cite{Wang2018Dataset} and data re-weighting\cite{pmlr-v80-ren18a}. Intuitively data augmentation aims to construct $G_{Inf}$ by checking trajectories of more data, while data distillation tries to shape $G_{Inf}$ with a minimal number of particle trajectories. The goal of data re-weighting is to construct a stable $G_{Inf}$ which is the critical point with respect to the re-weighting of data so that the landscape will be stable against the perturbation of data weights. How the training data can shape $G_{Inf}$ and what's the relationship between the training data size and the complexity of the deep network are interesting problems to be further explored in the furture.

\textbf{E. Attention, GAN, interaction and fibre bundle }
Attention mechanism became one of the most important components in deep learning systems in recent years. Following the guideline of geometrization of deep networks, it's interesting to examine the geometric and physical pictures of attention mechanism. Naturally the first thought can be, a deep network using attention mechanism is still a deep network so that its geometry has nothing special. Is this true or can we find a new geometry from attention mechanism?

The key idea of attention is to re-weight the information by coupling a normal deep network with a network that implements the attention mechanism. Intuitively if we take the deep network without attention as a curve in the manifold of the corner of physical functions, then coupling the attention module with the original deep network will shift the curve of the original network. This can be understood as that the attention mechanism applies a force on a particle along the curve so that the trajectory of the particle is shifted. So the attention mechanism is actually a mechanism to provide an interaction between data information. In physics interactions are usually geometrically represented by connections on fibre bundles. A geometric structure with connections and fibre bundles is an extension of the Riemannian structure. Similarly more complex deep networks such as the neural Turing machine(NTM) and the differential neural computer(DNC) can also be regarded as extended attention mechanisms. In our former work we have established a fibre bundle structure for the disentangled representation of deep networks\cite{Dong2019gauge}. How to build geometric structures of deep networks beyond Riemannian structure will also be our future work. 

Another related system is GAN, which is also a coupled deep network including the generator and the discriminator. How can GAN be formulated as a fibre bundle based interaction picture is an interesting issue. Another interesting question is how the structure of the generator of GAN can encode the data pattern, for example the pattern of images of human faces? Intuitively for a  generator to generate human face images, it can be understood as a cascaded error correcting code. During the generation of human face images, the generator not only encodes the input latent variable into the code words of human faces but also continuously corrects pattern errors so that perturbations or errors can be projected out of the code word subspace, just as shown in the work of \cite{Bau2018GAN} where wrong positions of church doors can be suppressed by the generator network. How can $G_{Inf}$ of the image generator encode an error correcting code in its geometry is an interesting topic to be explored in the future.

\textbf{F. Deep learning, meta learning and geodesics}
From the Riemannian structure $G_{Com}$ and the least complexity principle, conceptually the normal deep learning procedure is to find a geodesic to connect the trivial identity transformation and the target transformation on the corner of physical functions\ref{fig-1}. While meta learning such as MAML is to learn the geodesic equation on the manifold, which is encoded in the meta parameters, so that geodesics correspond to new tasks can be generated from the learned meta parameters given the target transformation encoded in the data of new tasks. Similarly, nonparametric methods are to learn the metric of $G_{Com}$ so that distances between transformations can be computed. The transformations for the test tasks can then be obtained by an interpolation of the training tasks based on the distances among them.

\section{Discussions}
The idea of geometry/information duality of deep networks can also be used to understand some interesting phenomena observed in a few recent papers.

\textbf{Weight agnostic neural network (WANN)}\\
In \cite{Gaier2019WANN} the so called weight agnostic neural networks(WANN) were introduced, whose function is more determined by the structure of a deep network instead of the weights. From the geometry/information view point, the encoded information of WANN is extremely robust with respect to the weights. From our discussion of the spacetime structure, we can see this can be understood as a kind of blackhole. This is an interesting observation since we can construct a black hole like structure in deep networks, which is a new evidence of the applicability of our programme of geometrization of deep networks. Here we need to point out that this is not a real black hole since real black holes have a more complex dynamics. We call WANN a black hole only because it has the same property of black holes that outside observer can not obtain a low variance estimate of its interior parameters. A straight forward question is, can we find worm holes in deep networks? From our understanding of the worm holes, they are related with quantum entanglement due to ER=EPR, so it's very unlikely that we can find worm holes in classical deep networks. But we can not exclude the possibility that we find something looks like a worm hole or some other interesting geometric structures. In \cite{Pastawski2015Holographic} Preskill modeled spacetime as a quantum error correcting code. In fact in the dissection of the image generator of GANs \cite{Bau2018GAN} we also see that the image generator network also works as a cascaded error correcting code so that it can encode the input latent vector into an image, while during this procedure possible error patterns can be corrected by the generator network. There is a high similarity between the generator network as a cascaded error correcting code and the QEC based spacetime structure. 

\textbf{Lottery ticket hypothesis}\\
Another recent hot spot about deep network is the lottery ticket hypothesis\cite{Frankle2018Ticket}, which essentially revealed the coupling between the structure and parameters of deep networks. From the geometry/information duality point of view, this observation is trivial since the geometry $G_{Inf}$ is determined by both the structure and the parameters so that they are coupled and only certain combinations of them can be used to represent the information of a certain application. What we are interested is the following up research\cite{zhou2019ticket} on the lottery hypothesis, where it claimed that ambitious information of the network parameters, such as the supermask or only the signs of the network parameters, can already encode partially the information. This means these ambitious information can already roughly shape the landscape of $G_{Inf}$. Obviously the robustness of the network function against the network configuration is closely related with the WANN case. How robust the geometry of $G_{Inf}$ is with respect to the network structure and parameters is also an interesting issue to be exploited in the future.

\emph{Overparameterized network and network pruning}
From the information/geometry duality point of view, overparameterized networks aim to build $G_{Inf}$ with an over-abundant number of nodes. On the contrary network pruning tries to approximate or interpolate $G_{Inf}$ of an overparameterized network with less points. Currently we see two different ways to achieve this goal. The dominating solution is to eliminate nodes with small weights step by step\cite{}. This strategy is very natural since nodes with small weights can be regarded as points with a small rise and fall in $G_{Inf}$ so that they can be first omitted.  Another idea is to first delete nodes staying in the geometric median of the configurations of all nodes\cite{He2019Median}. Geometrically such geometric medians are points that stay in the center of a set of nearby points so that theoretically such points can be interpolated by its neighbours. So both of these two strategies are geometrically valid. But which one is better? To see this, we can still go back to our geometric picture. If we regard the procedure of network pruning as a diffeomorphic transformation to match the initial overparameterized $G_{Inf}$ and the final simplified $G_{Inf}$ achieved by the pruned network, then the network pruning leads to a continuous curve connecting the two similar $G_{Inf}$s. This is an analogue of the quantum evolution between two quantum states. According to our understanding of the geometry of quantum computation, the optimal strategy to achieve the evolution is the one with the minimal energy spending. Intuitively the evolution trajectory should be smooth but not in a zig-zag pattern. From this point of view, of two different trajectories of the network pruning, we still prefer the trajectory to first delete nodes with small weights since it will deliver a smoother trajectory. On the other hand, if the strategy to delete geometric centers can converge, then this means the information saved in $G_{Inf}$ is robust to the configuration of the network so that the resulting network may have a better generalization performance.

\section{Conclusions}
Geometrization is not only the key idea of physics, it's also a framework to understand deep networks. Recently in physics research, the concept of computational complexity becomes a key player in understanding the structure of spacetime and quantum phase. In this work, we formally propose the programme of geometrization of deep networks as a framework for the interpretability of deep learning systems. Inspired by the information/geometry duality in physics, we examined the geometry/information duality in deep networks so that we can build two geometric structures on deep networks. As an analogue of the geometric mechanics or the geometry of quantum computation, the first geometric structure $G_{Com}$ defined a Riemannian manifold on the space of the functions that can be represented by deep networks, where the Riemannian metric plays a role to define the complexity of deep networks. When this Riemannian metric is chosen as the Fisher information metric, which aims to represent how the configuration of a deep network is related with the information represented by the deep network, a second geometric structure $G_{Inf}$ emerges, which can be regarded as a memory spacetime that can encode the information by its geometric structure.

In this geometric picture of deep networks, the Fisher information metric plays a central role to integrate the two geometric structures, $G_{Com}$ and $G_{Inf}$. On one hand it defines a Riemannian metric to measure the network complexity. On the other hand, the complexity metric aims to optimize the way how the deep network encodes the information of data. Applying a least action principle on the Fisher information metric based network complexity results in not only an optimal deep network with a minimal network complexity, which is a geodesic on the Riemannian manifold $G_{Com}$ connecting the identity function and the target function that can accomplish the task, but also a second geometric structure $G_{Inf}$ which can encode the information of data in its geometry with a high efficiency and robustness.

This geometric picture of deep networks can help us to have a general framework to analysis the properties of deep networks including the network complexity, generalization, disentangled representation and other related issues.  

We hope this observation can serve as a strong evidence that our programme of geometrization of deep networks is not only a promising framework for the interpretability of deep learning systems, but also it can bridge the concepts of deep networks and physics so that it also provide the interpretability of our physical world.

\bibliographystyle{unsrt}

\bibliography{DLmemory}

\begin{thebibliography}{10}

\bibitem{Dong2019geo}
X.~Dong and L.~Zhou.
\newblock Geometrization of deep networks for the interpretability of deep
  learning systems.
\newblock {\em arxiv:1901.02354}, 2019.

\bibitem{Dong2019over}
X.~Dong and L.~Zhou.
\newblock Understanding over-parameterized deep networks by geometrization.
\newblock {\em arxiv:1902.03793}, 2019.

\bibitem{Lloyd2006A}
S.~Lloyd.
\newblock A theory of quantum gravity based on quantum computation.
\newblock {\em Class.quant.grav}, 2006.

\bibitem{Swingle2012Constructing}
Brian Swingle.
\newblock Constructing holographic spacetimes using entanglement
  renormalization.
\newblock {\em Physics}, 2012.

\bibitem{Raamsdonk2010Building}
M.~van Raamsdonk.
\newblock Building up spacetime with quantum entanglement.
\newblock {\em General Relativity and Gravitation}, 42(10):2323--2329, 2010.

\bibitem{Matsueda2014Derivation}
Hiroaki Matsueda.
\newblock Derivation of gravitational field equation from entanglement entropy.
\newblock {\em arXiv:1408.5589v2}, 70, 2014.

\bibitem{Wen2003Quantum}
Wen Xiao-Gang.
\newblock Quantum order from string-net condensations and the origin of light
  and massless fermions.
\newblock {\em Physical Review D}, 68(6):484--504, 2003.

\bibitem{Lloyd2012A}
Seth Lloyd.
\newblock A theory of quantum gravity based on quantum computation.
\newblock {\em Class.quant.grav}, 2012.

\bibitem{Evenbly2011Tensor}
G.~Evenbly and G.~Vidal.
\newblock Tensor network states and geometry.
\newblock {\em Journal of Statistical Physics}, 145(4):891--918, 2011.

\bibitem{Evenbly2017Algorithms}
Glen Evenbly.
\newblock Algorithms for tensor network renormalization.
\newblock {\em Phys.rev.b}, 95(4), 2017.

\bibitem{Nielsen_geometry2}
M.~R. Dowling and M.~A. Nielsen.
\newblock The geometry of quantum computation.
\newblock {\em Quantum Information and Computation}, 8(10):861--899, 2008.

\bibitem{Susskind2017Dear}
Leonard Susskind.
\newblock Dear qubitzers, gr=qm.
\newblock {\em arXiv:1708.03040v1}, 2017.

\bibitem{Dong_deep}
J.S.~Wu X.~Dong and L.~Zhou.
\newblock How deep learning works --the geometry of deep learning.
\newblock {\em arXiv:1710.10784}, 2017.

\bibitem{Bruveris2013Geometry}
Martins Bruveris and Darryl~D. Holm.
\newblock Geometry of image registration: The diffeomorphism group and momentum
  maps.
\newblock {\em Fields Institute Communications}, 73:19--56, 2013.

\bibitem{Weinan2018A}
E~Weinan, Jiequn Han, and Qianxiao Li.
\newblock A mean-field optimal control formulation of deep learning.
\newblock {\em arxiv:1807.01083v1}, 2018.

\bibitem{Or2014A}
Román Orús.
\newblock A practical introduction to tensor networks: Matrix product states
  and projected entangled pair states.
\newblock {\em Annals of Physics}, 349(10):117--158, 2014.

\bibitem{Susskind_ER_bridge}
L.~Susskind.
\newblock Entanglement is not enough.
\newblock {\em arXiv:1411.0690v1}, 2014.

\bibitem{Matsueda2013Emergent}
Hiroaki Matsueda.
\newblock Emergent general relativity from fisher information metric.
\newblock {\em arXiv:1310.1831v2}, 2013.

\bibitem{Bruveris2016Geometry}
Martins Bruveris and Peter~W. Michor.
\newblock Geometry of the fisher-rao metric on the space of smooth densities on
  a compact manifold.
\newblock 2016.

\bibitem{pmlr-v70-koh17a}
Pang~Wei Koh and Percy Liang.
\newblock Understanding black-box predictions via influence functions.
\newblock In Doina Precup and Yee~Whye Teh, editors, {\em Proceedings of the
  34th International Conference on Machine Learning}, volume~70 of {\em
  Proceedings of Machine Learning Research}, pages 1885--1894, International
  Convention Centre, Sydney, Australia, 06--11 Aug 2017. PMLR.

\bibitem{pmlr-v80-ren18a}
Mengye Ren, Wenyuan Zeng, Bin Yang, and Raquel Urtasun.
\newblock Learning to reweight examples for robust deep learning.
\newblock In Jennifer Dy and Andreas Krause, editors, {\em Proceedings of the
  35th International Conference on Machine Learning}, volume~80 of {\em
  Proceedings of Machine Learning Research}, pages 4334--4343,
  Stockholmsmässan, Stockholm Sweden, 10--15 Jul 2018. PMLR.

\bibitem{Liang2017Fisher}
Tengyuan Liang, Tomaso Poggio, Alexander Rakhlin, and James Stokes.
\newblock Fisher-rao metric, geometry, and complexity of neural networks.
\newblock {\em arxiv:1711.01530}, 2017.

\bibitem{Lei2017Towards}
Wu~Lei, Zhanxing Zhu, and E~Weinan.
\newblock Towards understanding generalization of deep learning: Perspective of
  loss landscapes.
\newblock {\em arxiv:1706.10239v2}, 2017.

\bibitem{Karakida2018Universal}
Ryo Karakida, Shotaro Akaho, and Shun~Ichi Amari.
\newblock Universal statistics of fisher information in deep neural networks:
  Mean field approach.
\newblock 2018.

\bibitem{Susskind2016The}
Leonard Susskind.
\newblock The typical state paradox: diagnosing horizons with complexity.
\newblock {\em Fortschritte Der Physik}, 64(1):84--91, 2016.

\bibitem{Susskind_ER_bridge_nowhere}
L.~Susskind and Y.~Zhao.
\newblock Switchbacks and the bridge to nowhere.
\newblock {\em arXiv:1408.2823v1}, 2014.

\bibitem{Fern2015Frustration}
C.~Fernández-González, N.~Schuch, M.~M. Wolf, J.~I. Cirac, and
  D.~Pérez-García.
\newblock Frustration free gapless hamiltonians for matrix product states.
\newblock {\em Communications in Mathematical Physics}, 333(1):299--333, 2015.

\bibitem{Heydari_dynamicdiatance}
H.~Heydari.
\newblock Geometric formulation of quantum mechanics.
\newblock {\em arXiv:1503.00238}, 2015.

\bibitem{Locatello2018Challenging}
Francesco Locatello, Stefan Bauer, Mario Lucic, Gunnar Rätsch, Sylvain Gelly,
  Bernhard Schölkopf, and Olivier Bachem.
\newblock Challenging common assumptions in the unsupervised learning of
  disentangled representations.
\newblock 2018.

\bibitem{Dong2019gauge}
X.~Dong and Zhou. L.
\newblock Gauge theory and twins paradox of disentangled representations.
\newblock {\em arxiv:1906.10545}, 2019.

\bibitem{Wang2018Dataset}
Tongzhou Wang, Jun-Yan Zhu, Antonio Torralba, and Alexei~A. Efros.
\newblock Dataset distillation.
\newblock 2018.

\bibitem{Bau2018GAN}
David Bau, Jun-Yan Zhu, Hendrik Strobelt, Bolei Zhou, Joshua~B. Tenenbaum,
  William~T. Freeman, and Antonio Torralba.
\newblock Gan dissection: Visualizing and understanding generative adversarial
  networks.
\newblock 2018.

\bibitem{Gaier2019WANN}
A.~Gaier and D.~Ha.
\newblock Weight agnostic neural networks.
\newblock {\em arxiv:1906.04358}, 2019.

\bibitem{Pastawski2015Holographic}
D.~Harlow F.~Pastawski, B.~Yoshida and J.~Preskill.
\newblock Holographic quantum error-correcting codes: toy models for the
  bulk/boundary correspondence.
\newblock {\em Journal of High Energy Physics}, 2015(6):1--55, 2015.

\bibitem{Frankle2018Ticket}
Jonathan Frankle and Michael Carbin.
\newblock The lottery ticket hypothesis: Finding sparse, trainable neural
  networks.
\newblock {\em arxiv:1803.03635}, 2018.

\bibitem{zhou2019ticket}
H.~Zhou, J.~Lan, R.~Liu, and J.~Yosinski.
\newblock Deconstructing lottery tickets: Zeros, signs, and the supermask.
\newblock {\em arxiv:1905.01067}, 2019.

\bibitem{He2019Median}
Y.~He, P.~Liu, Z.W. Wang, Z.L. Hu, and Yang Y.
\newblock Filter pruning via geometric median for deep convolutional neural
  networks acceleration.
\newblock {\em arxiv:1811.00250}, 2018.

\end{thebibliography}



%\vfill


\end{document}